\documentclass[sigconf]{acmart}

\AtBeginDocument{%
  }
    
\usepackage{amsmath}
\usepackage{amsfonts}
\usepackage{algorithm}
\usepackage{algorithmic}
\usepackage{booktabs}
\usepackage{graphicx,xcolor}
\usepackage{hyperref}
\usepackage{float}
\usepackage{xspace}

\newtheorem{property}{Property}

\AtBeginDocument{%
  }
    
\copyrightyear{2022} 
\acmYear{2022} 
\setcopyright{acmcopyright}\acmConference[MM '22]{Proceedings of the 30th ACM International Conference on Multimedia}{October 10--14, 2022}{Lisboa, Portugal}
\acmBooktitle{Proceedings of the 30th ACM International Conference on Multimedia (MM '22), October 10--14, 2022, Lisboa, Portugal}
\acmPrice{15.00}
\acmDOI{10.1145/3503161.3548110}
\acmISBN{978-1-4503-9203-7/22/10}

\begin{document}

\title{On Generating Identifiable Virtual Faces}

\author{Zhuowen Yuan}
\affiliation{%
  \institution{Fudan University}
  \city{}
  \country{}}
\email{zwyuan18@fudan.edu.cn}

\author{Zhengxin You}
\affiliation{%
  \institution{Fudan University}
  \city{}
  \country{}}
\email{zxyou20@fudan.edu.cn}

\author{Sheng Li}
\authornote{Corresponding author}
\affiliation{%
  \institution{Fudan University}
  \city{}
  \country{}}
\email{lisheng@fudan.edu.cn}

\author{Zhenxing Qian}
\authornotemark[1]
\affiliation{%
  \institution{Fudan University}
  \city{}
  \country{}}
\email{zxqian@fudan.edu.cn}

\author{Xinpeng Zhang}
\affiliation{%
  \institution{Fudan University}
  \city{}
  \country{}}
\email{zhangxinpeng@fudan.edu.cn}

\author{Alex C. Kot}
\affiliation{%
  \institution{Nanyang Technological University}
  \city{}
  \country{}}
\email{eackot@ntu.edu.sg}

\renewcommand{\shortauthors}{Zhuowen Yuan et al.}

\begin{abstract}
Face anonymization with generative models have become increasingly prevalent since they sanitize private information by generating virtual face images, ensuring both privacy and image utility. Such virtual face images are usually not identiﬁable after the removal or protection of the original identity. In this paper, we formalize and tackle the problem of generating identiﬁable virtual face images. Our virtual face images are visually different from the original ones for privacy protection. In addition, they are bound with new virtual identities, which can be directly used for face recognition. We propose an Identiﬁable Virtual Face Generator (IVFG) to generate the virtual face images. The IVFG projects the latent vectors of the original face images into virtual ones according to a user speciﬁc key, based on which the virtual face images are generated. To make the virtual face images identiﬁable, we propose a multi-task learning objective as well as a triplet styled training strategy to learn the IVFG. We evaluate the performance of our virtual face images using different face recognizers on diffident face image datasets, all of which demonstrate the effectiveness of the IVFG for generate identiﬁable virtual face images.
\end{abstract}


\maketitle

\section{Introduction}

Since the advent of the digital era, millions of images are shared over the Internet with the help of personal devices.
Although they enable us to build stronger connections with the society, there is growing concern that these techniques may introduce privacy issues.
Often, the images that are uploaded without sanitization contain privacy-sensitive information of individuals, such as identity, location, etc.
Such private information may be leveraged by adversaries for malicious purposes.
The public available images may also be collected and utilized to train machine learning models without consent \cite{harvey2019megapixels}.

To tackle these problems, obfuscation techniques such as pixelization and blurring \cite{hill2016effectiveness} are widely adopted. However, these traditional methods fail to protect against modern attacks based on deep neural networks. McPherson \emph{et al.} \cite{mcpherson2016defeating} show that a trained attack model can re-identify the obfuscated face images with over 95\% of accuracy. On the other hand, the quality of these obfuscated images are severely distorted, which could hardly be used for other computer vision tasks. Recently, researchers propose to add adversarial noise to the face images \cite{shan2020fawkes, raval2017protecting} to mislead the face recognizer. These models rely on the prior knowledge of the face recognizer for high performance. People can easily reveal the original identity of the obfuscated face image by human perception \cite{li2021deepblur}.

A growing trend of researches take advantage of generative adversarial networks (GANs) \cite{goodfellow2014generative} for generating realistic face images from original face images \cite{maximov2020ciagan, mirjalili2020privacynet, li2021deepblur, kuang21,gu2020password}. One branch of these methods performs attribute editing while preserving the original identity. For instance, Mirjalili \emph{et al.} \cite{mirjalili2020privacynet} propose to remove sensitive soft attributes so that the generated faces can be used for face authentication, but the facial attributes can no longer be classified. Such a scheme achieves a low level of face private protection as the original identity is still revealed in the anonymized face image. 

\begin{figure*}[t]
    \centering
    \includegraphics[width=0.8\textwidth]{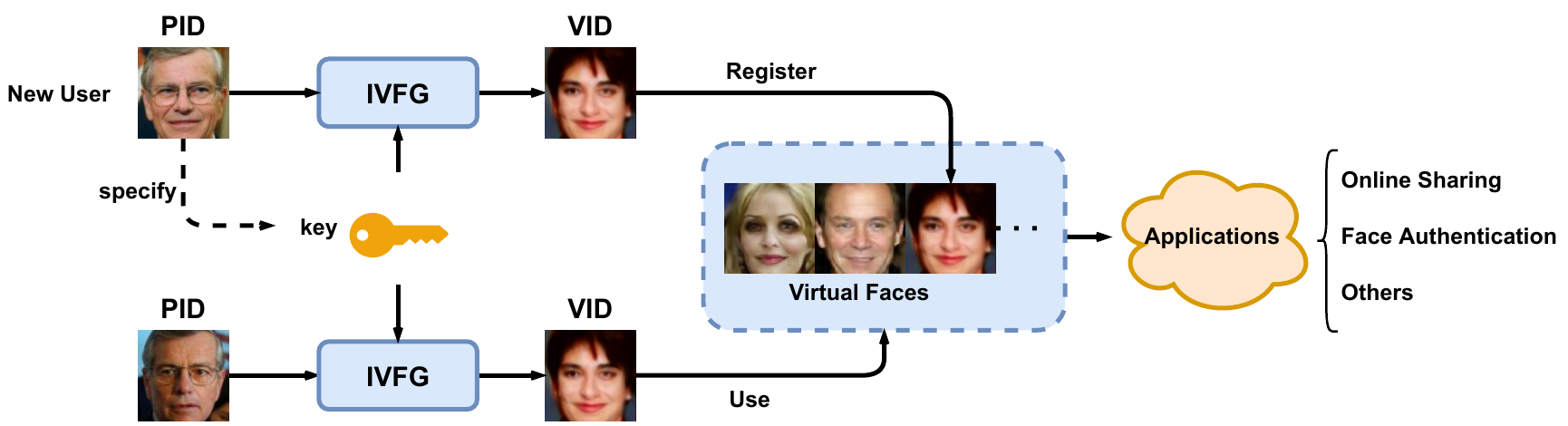}
    \caption{A typical application of our IVFG framework. Suppose that a user does not want to send his original face images to the untrusted applications, but still wishes to be identified. He can generate a virtual face image with our framework locally, and then feed the faces of his unique VID to the untrusted applications. Downstream applications are all performed in the virtual space, thus protecting the privacy of PIDs.}
    \label{fig1}
\end{figure*}

Another category anonymizes the identity of the face image, where the original identity is not revealed in the generated face images. Various face image anonymization networks have been proposed, including the deepblur network \cite{li2021deepblur}, the conditional identity anonymization GAN \cite{maximov2020ciagan}, and the de-identification generative adversarial network (DeIdGAN) \cite{kuang21}. These approaches are able to generate anonymized face images with relatively good resuability in terms of visual quality and face detection. However, the identity information in the original face images are removed forever, which can not be used for authentication when needed.

Recently, attempts have been made to encrypt the original face images into real look alike anonymized face images \cite{gu2020password}. Different anonymized face images could be generated from the original face images using different keys. In addition, the original face images can be recovered from the anonymized face images with the correct key available. As such, face authentication can be carried out in the original space after decryption. However, the original identity is leaked to the third party who performs face authentication.


With the widespread applications of face identification techniques, it is urgent to develop countermeasures against illegal usage of face images. Unlike other images, face images contain unique identity information, which can not be replaced or reissued. Once stolen, people's identities will be compromised forever. Therefore, it is important that the original identity can not be revealed from the anonymized face images. On the other hand, the collection of face images is expensive and requires a lot of human effort. Thus, the resuability should not be compromised after anonymization, while the most important usability of a face image is the identifiability. 

In this paper, we try to tackle the aforementioned issues by generating identifiable virtual face images for privacy protection. A virtual face image is generated from the original face image according to a key, which satisfies the following requirements.
\begin{itemize}
    \item The virtual face images are visually different from the original face images for anonymization.
    \item Different virtual face images can be generated from the same original face image with different keys. 
    \item The virtual face images can be directly used for face authentication in the virtual space.
\end{itemize}

Concretely, we propose a novel Identifiable Virtual Face Generator (IVFG) to fulfill such requirements. The IVFG maps an original face image into different virtual face images using different keys. For face images from the same original identity, the IVFG is able to generate virtual face images belonging to a unique virtual identity (VID) using a user specific key. In the following discussions, we also term the original identity as the physical identity (PID). The virtual face images can be used in different applications that require the face images as inputs for anonymization purpose. Take the online face authentication as an example, we can submit the virtual faces and only the VID is used for registration or authentication. Thus, the face authentication can be conducted in the virtual space to protect the user’s privacy, where the PID is never exposed online as shown in Fig. \ref{fig1}. 

Our proposed IVFG framework is based on an Encoder-Decoder architecture. We propose to project the features of the original face images to the latent space of the Style-GAN \cite{karras2019style} according to a user specific key. This is to encode the VID information into the virtual face images which are the outputs of the Style-GAN decoder. To make the virtual face images identifiable, a multi-task learning objective is further proposed to achieve the following properties, including the anonymization, diversity, identifiability and separability. The performance of our virtual faces are shown to be promising using different face recognizers on different face image datasets. 

The main contributions are summarized below.
\begin{itemize}
    \item We first explore the possibility to generate identifiable virtual face images which protects the PID and can be identified directly in the virtual space with off-the-shelf face recognizers.
	\item We propose a novel framework IVFG to map PIDs to VIDs conditioned on a user-specified key.
	\item We propose a multi-task learning objective as well as the corresponding training strategy to facilitate the training of IVFG.
\end{itemize}

\section{Related Work}
In this section, we brieﬂy summarize related work on image obfuscation, GANs, and face anonymization.

\subsection{Image Obfuscation}

Traditional methods such as pixelization, blurring \cite{hill2016effectiveness} are widely adopted for image obfuscation. These methods severely reduce the image quality and utility. They also fail to protect private information in images against deep models as already demonstrated by \cite{mcpherson2016defeating}. Another category of methods leverage the power of adversarial machine learning \cite{kurakin2017adversarial, shan2020fawkes,raval2017protecting} to fool the classiﬁer. Shan \emph{et al.} \cite{shan2020fawkes} propose to add cloaks (i.e., adversarial noise) to the images. It achieves over 95 \% protection rate across different feature extractors. However, such obfuscated images are still recognizable by human perception \cite{li2021deepblur}. Raval \emph{et al.} \cite{raval2017protecting} train an adversarial network by jointly optimizing the privacy and utility objectives to perturb the images for privacy protection. This method relies on the prior knowledge of the attacker’s network, which may not be applicable in many real-world scenarios.

\subsection{Generative Adversarial Networks}
The idea of GAN is initially introduced by \cite{goodfellow2014generative}, which has widely been used for image generation. A typical GAN consists of a discriminator and a generator, which are jointly trained in a minimax game. The discriminator learns to distinguish between real and fake faces while the generator learns to generate realistic samples to fool the discriminator. Karras \emph{et al.} \cite{karras2019style} propose a style-based generator StyleGAN and achieve good performance. Developments in GANs have also enabled controllable generation given input labels \cite{mirza2014conditional}. Abdal \emph{et al.} \cite{abdal2021styleflow} investigate the latent space of StyleGAN and propose StyleFlow for attribute-conditioned semantic editing on projected real images. Richardson \emph{et al.} \cite{richardson2021encoding} propose an encoder for StyleGAN inversion, which demonstrates the possibility of face translation using pre-trained models. Various optimization strategies have been proposed \cite{arjovsky2017wasserstein, mao2017least} to facilitate the training of GANs.

\vspace{-10pt}
\subsection{Face Anonymization}

Face anonymization refers to publishing face images with the identity information sanitized. The image obfuscation techniques mentioned earlier can also be utilized for face anonymization. Recent advances take advantage of GAN to generate sanitized face images \cite{yan2019attributes,mirjalili2020privacynet}. Li and Lin \cite{li2019anonymousnet} propose AnonymousNet that performs face attribute editing based on a set of privacy metrics. Similarly, Mirjalili \emph{et al.} \cite{mirjalili2020privacynet} propose to generate anonymized face images by removing sensitive soft attributes. These schemes only protect the privacy of the face attributes, the original identities (i.e., the PIDs) are preserved in the anonymized face images for authentication.

In order to protect the PIDs of the face images, Li and Choi \cite{li2021deepblur} propose DeepBlur to perform blurring in the latent space of a pre-trained generative model, which is capable of synthesizing realistic face images with the PIDs removed. Maximov \emph{et al.} \cite{maximov2020ciagan} propose a general face anonymization framework by mixing styles of different identities using GAN. Kuang \emph{et al.} \cite{kuang21} propose an de-identification generative adversarial network (DeIdGAN) for face anonymization, which seamlessly replaces the original face image by a visually different but realistic face image. The anonymized face images generated by such schemes can only be used for face detection, which are not able to be utilized for authentication anymore due to the removal of PIDs. To deal with this issue, Gu \emph{et al.} \cite{gu2020password} propose a face image encryption framework to generate visually realistic encrypted face images bases on different passwords. The encrypted face images are generated based on a password conditioned GAN, which well protect the PIDs and can be decrypted into the original face images for authentication. Despite the advantage, this method fails to protect the PIDs after the face image decryption. The protection mainly relies on the secrecy of the password, and the anonymized face images can not be used directly for face authentication.


In this paper, we propose to generate identiﬁable virtual face images for anonymization. The virtual face images are generated based on both the original face image and a user speciﬁc key. Unlike the existing face anonymization schemes, our virtual faces can be authenticated directly in the virtual space without leaking the PIDs.

\section{Problem Formulation}

Let $\mathcal{X}$ and $\mathcal{Y}$ denote two sets of face images, where images within each set belong to the same person (i.e., the same PID). The anonymization process is conditioned on a user-specified key $k \in \mathcal{K}$.
We further denote $T(x, k)$ as the virtual face image generated from the original face image $x$ and key $k$, and $I(x)$ as the identity that can be extracted from $x$.

Given the original face images $x_1, x_2 \in \mathcal{X}$, $y \in \mathcal{Y}$ and two disparate keys $k_1, k_2 \in \mathcal{K}$, our goal is to generate virtual face images that meet the following properties.

\begin{property}[Anonymization]
    The original face image and virtual face image should belong to different identities, namely VID $\neq$ PID:
    \label{prop:anony}
\end{property}
\begin{equation}
    I(T(x_1, k_1)) \neq I(x_1).
\end{equation}

\begin{property}[Diversity]
    The key controls the VID of the anonymized face images, i.e., the virtual face images of the same PID belong to different VIDs under different keys:
    \label{prop:diversity}
\end{property}
\begin{equation}
    I(T(x_1, k_1)) \neq I(T(x_1, k_2)).
\end{equation}

\begin{property}[Identifiability]
    Given two original face images belonging to the same PID, the corresponding virtual face images should belong to the same VID under the same key:
    \label{prop:homo}
\end{property}
\begin{equation}
    I(T(x_1, k_1)) = I(T(x_2, k_1)).
    \label{eq1:prop3}
\end{equation}

\begin{property}[Separability]
    Given two original face images belonging to different PIDs, the corresponding virtual face images should belong to different VIDs under the same or different keys:
    \label{prop:robust}
\end{property}
\begin{equation}
\begin{split}
    I(T(x_1, k_1)) & \neq I(T(y, k_1)) \\
    I(T(x_1, k_1)) & \neq I(T(y, k_2)).
    \label{eq2:prop3}
\end{split}
\end{equation}


\begin{figure*}[t]
    \centering
    \includegraphics[width=0.8\textwidth]{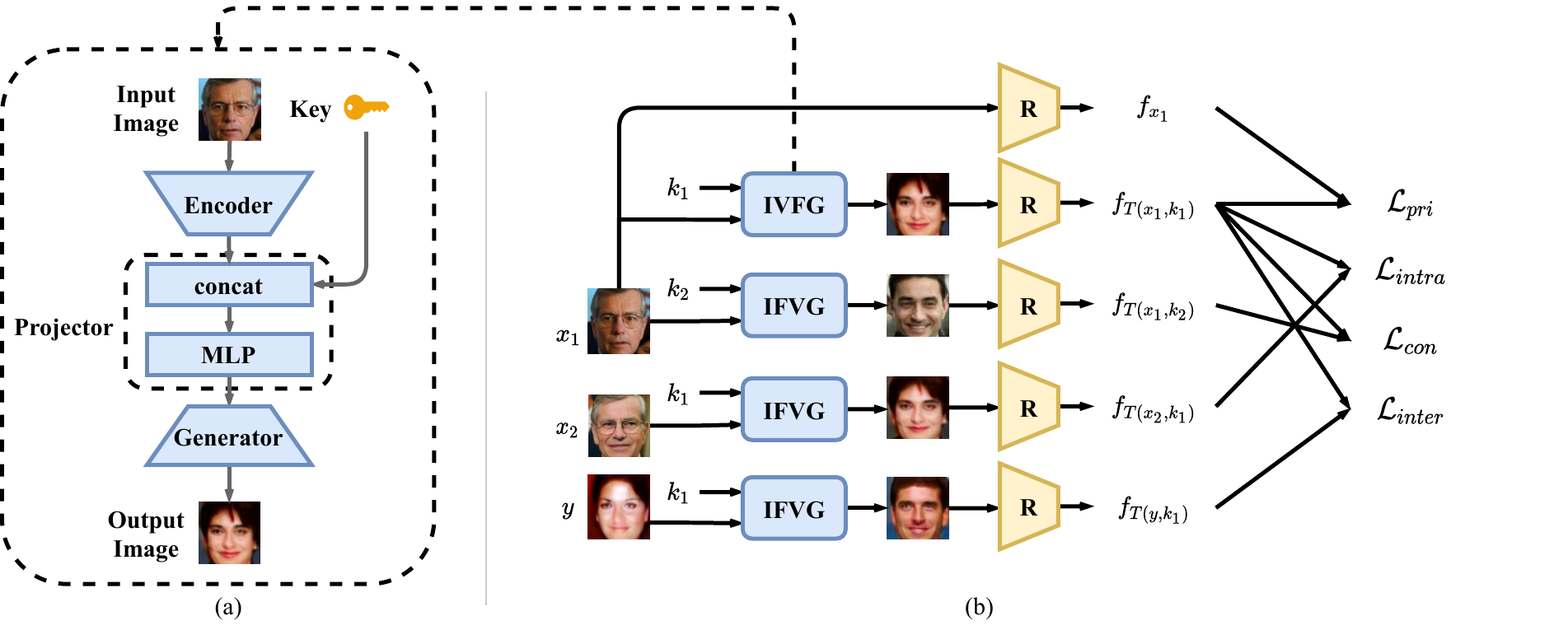}
    \caption{(a) The proposed IVFG framework. (b) Overview of our training strategy.}
    \label{fig3}
\end{figure*}

\section{Methodology}
In this section, we introduce our IVFG framework, learning objective and training strategy.

\subsection{The Proposed Framework}
The overall structure of our proposed IVFG framework is illustrated in Fig. \ref{fig3}(a). The IVFG framework is based on an encoder-decoder architecture, which is composed of three modules: an encoder $E$, a latent projector $P$ and a face generator $G$. The encoder $E$ takes an image tensor $x$ as input and outputs the extracted facial representation, say $r$. We propose a latent projector $P$ to project the combination of facial representation $r$ and key $k$ to the latent space of the generator, which is composed of a concatenation operation and a multilayer perceptron (MLP). We use the generator of a pre-trained GAN as our $G$, which maps a vector in latent space $z \in \mathcal{Z}$ to realistic images from the distribution of training data $x \sim p(x)$, namely: $G: z \mapsto x$.

\subsection{Learning Objective}
We design a multi-task learning objective to satisfy the four properties mentioned in Section 3 for training the IVFG. We leverage the following cosine embedding loss $\mathcal{L}_{emb}$ to measure the similarity between two facial features $f_1$ and $f_2$, where
\begin{equation}
    \mathcal{L}_{emb}(f_1, f_2, l) = \\
    \begin{cases}
        1 - \cos(f_1, f_2), & l = 1 \\
        \max(m, \cos(f_1, f_2)), & l = -1 \\
    \end{cases},
\end{equation}
where $cos$ denotes cosine similarity, $m$ is a hyperparameter, and $l$ is the identity indicator. $l=1$ refers to the case that $f_1$ and $f_2$ to be from the same identity, and we are pushing $f_1$ and $f_2$ to be close to each other by optimizing $\mathcal{L}_{emb}$. On the contrary, we try to separate $f_1$ and $f_2$ to make them belong to different identities if $l$ is set to $-1$.

\textit{Privacy Loss.}
To satisfy Property \ref{prop:anony}, we propose privacy loss $\mathcal{L}_{pri}$ to modify the identity of the original face image. In other words, $\mathcal{L}_{pri}$ guides the anonymization process so that PID $\neq$ VID, where
\begin{equation}
    \mathcal{L}_{pri} = \mathbb{E}_{\mathcal{X}, \mathcal{K}}[\mathcal{L}_{emb}(R(T(x_1,k_1)), R(x_1), -1)],
\end{equation}
where $R(x)$ extracts the features from the face image $x$.

\textit{Conditional Loss.}
To satisfy Property \ref{prop:diversity}, we propose conditional loss $\mathcal{L}_{con}$ to make sure that the generated virtual identity can be controlled by a user-specified key. Given two different keys $k_1, k_2$ and an original face image, the model should be able to generate two distinct VIDs by using $\mathcal{L}_{con}$ below
\begin{equation}
    \mathcal{L}_{con} = \mathbb{E}_{\mathcal{X}, \mathcal{K}}[\mathcal{L}_{emb}(R(T(x_1, k_1)), R(T(x_1,k_2)), -1)].
\end{equation}

\textit{Intra-group Classification Loss.}
To satisfy Property \ref{prop:homo}, we have to guarantee that, given two original face images from the same PID: $x_1, x_2 \in \mathcal{X}$, the corresponding virtual face images generated using same key should belong to the same VID. Therefore, the intra-group classification loss is formulated as
\begin{equation}
    \mathcal{L}_{intra} = \mathbb{E}_{\mathcal{X}, \mathcal{K}}[\mathcal{L}_{emb}(R(T(x_1,k_1)), R(T(x_2,k_1)), 1)].
\end{equation}

\textit{Inter-group Classification Loss.}
To satisfy Property \ref{prop:robust}, we propose the following inter-group classification loss $\mathcal{L}_{inter}$ to maximize the distance between VIDs that are generated from different PIDs
\begin{equation}
    \mathcal{L}_{inter} = \mathbb{E}_{\mathcal{X}, \mathcal{Y}, \mathcal{K}}[\mathcal{L}_{emb}(R(T(x_1,k_1)), R(T(y,k_1)), -1)],
\end{equation}
where $x$ and $y$ are face images from different PIDs.

\noindent \textit{Regularization.}
With the above losses, the predicted latent vectors may deviate from the latent space of $G$, leading to non-convergence. To deal with this issue, we further propose a regularization term to bound the distribution of the projected latent vectors such that they are within the latent space of $G$. Specifically, we first calculate the mean latent vector $\bar{z}$ by sampling \emph{m} latent vectors $z_1, z_2, ..., z_m$ from the latent distribution of $G$. Then, we minimize the L2 distance between the predicted latent vector $z$ and $\bar{z}$ by
\begin{equation}
    \mathcal{L}_{reg} = ||z - \bar{z}||_2^2.
\end{equation}

Our full objective is a weighted average over the above mentioned loss functions, which is given below
\begin{equation}
\begin{split}
    \mathcal{L}&=\lambda_{pri}\mathcal{L}_{pri}+\lambda_{con}\mathcal{L}_{con}\\
    &+\lambda_{intra}\mathcal{L}_{intra}+\lambda_{inter}\mathcal{L}_{inter}+\lambda_{reg}\mathcal{L}_{reg}.
\end{split}
\label{eq:full}
\end{equation}

\subsection{Training Strategy}
During training, we leverage an auxiliary face recognizer $R$ for facial feature extraction. We use pre-trained models for $E$, $G$, $R$ whose parameters will be frozen in the training. We only update the parameters of $P$ by
\begin{equation}
    P^* = \arg \min_{P}{\mathcal{L}}.
\end{equation}

We split the dataset $\mathcal{S}$ into triplets of images. For each triplet $s \in \mathcal{S}$, we have $s=\{x_1, x_2, y\}$ with $x_1, x_2$ belonging to the same PID and $y$ for another PID. Given two different keys $k_1$ and $k_2$, we can derive a quaternion $s'$ of four virtual images: $T(x_1,k_1)$, $T(x_1,k_2)$, $T(x_2,k_1)$, $T(y,k_1)$. Then, we calculate the objective function and perform back propagation. Such a triplet styled training strategy is illustrated in Fig. \ref{fig3}(b) with the details given in Algorithm \ref{alg1}.

\begin{algorithm}[tb]
    \caption{IVFG Training}
    \label{alg1}
    \textbf{Input}: $E$: pre-trained encoder; $G$: pre-trained generator; $R$: pre-trained face recognizer; $P(f, k; W)$: latent projector; $W$: parameters of $P$, $\mathcal{S}_n$: dataset containing $n$ triplets\\
    \textbf{Parameter}: $\lambda_{pri}, \lambda_{con}, \lambda_{intra}, \lambda_{inter}, \lambda_{reg}$: weights for the training objective; $\alpha$: learning rate
    \begin{algorithmic}[1]
    \STATE Freeze $E$, $G$, $R$.
    \STATE // Generate $\hat z$ by sampling from latent space.
    \STATE $z_1, ..., z_m = sample(z), z \sim p(z)$
    \STATE $\hat z = \frac{1}{m} \sum_{j=1}^{m}z_j$
    \FOR {$i \gets 1$ to $n$}
    \STATE $k_1, k_2 \gets$ generate two distinct keys
    \STATE $x_1, x_2, y \gets \mathcal{S}_i$
    \STATE $r_{x_1} \gets E(x_1)$ // Extract facial representations.
    \STATE $r_{x_2} \gets E(x_2)$
    \STATE $r_{y} \gets E(y)$
    \STATE $z_{x_1,k_1} \gets P(r_{x_1}, k_1; W_i)$ // Project to latent vectors.
    \STATE $z_{x_1,k_2} \gets P(r_{x_1}, k_2; W_i)$
    \STATE $z_{x_2,k_1} \gets P(r_{x_2}, k_1; W_i)$
    \STATE $z_{y,k_1} \gets P(r_{y}, k_1; W_i)$
    \STATE $T(x_1,k_1) \gets G(z_{x_1,k_1})$ // Generate virtual faces.
    \STATE $T(x_1,k_2) \gets G(z_{x_1,k_2})$
    \STATE $T(x_2,k_1) \gets G(z_{x_2,k_1})$
    \STATE $T(y,k_1) \gets G(z_{y,k_1})$
    \STATE Calculate full objective $\mathcal{L}$ from Eq. \ref{eq:full}.
    \STATE $\nabla W_i = \partial \mathcal{L}/\partial W_i$
    \STATE $W_{i+1} = W_i - \alpha \nabla W_i$
    \ENDFOR
    \STATE \textbf{return} $W_{i+1}$
    \end{algorithmic}
\end{algorithm}

\section{Experiments}

\subsection{Datasets}
Our model is trained on LFW \cite{LFWTech} and evaluated on both LFW \cite{LFWTech} and CelebA \cite{liu2015faceattributes}. The LFW dataset consists of 13,233 face images from 5,749 identities, including 1,680 identities with two or more images. We perform an 8:1:1 split based on the identities for training, validation, and testing. Then, we split each set into triplets of images as discussed in the previous section. We also randomly select 10\% of the face images in the CelebA dataset as another test set for cross dataset evaluation. All face images are aligned and cropped to $128 \times128$ by MTCNN \cite{zhang2016joint}.

\subsection{Implementation Details and Settings}
We use FaceNet Inception-ResNet-v1 \cite{szegedy2017inception} for both $E$ and $R$ in training.
Both of them are pre-trained on VGGFace2 \cite{cao2018vggface2}.
For the input key, we use an $n$-digit binary string, which is represented by a $n$-dimensional vector $k \in \{0, 1\}^n$.
We set $n=8$ as the default value.
$G$ is a StyleGAN \cite{karras2019style} generator trained from scratch on LFW with progressive growing \cite{karras2018progressive}.
We stop after the 128-pixel training step since we only use the model for generating $128 \times128$ images \footnote{If a higher resolution is required, we can pre-train the StyleGAN for more iterations with progressive growing until the generator learns how to generate high resolution images.}.
For effective and stable evaluation, all noises included in the original StyleGAN implementation are removed during both training and evaluating.
Our projector $P$ maps the concatenation of facial representation and key into the intermediate latent space $\mathcal{W}$ of StyleGAN instead of $Z$, since $\mathcal{W}$ achieves better disentanglement as pointed out by \cite{karras2019style}.
$w \in \mathcal{W}$ directly controls the generator through adaptive instance normalization \cite{huang2017arbitrary} at each convolution layer.

We set the number of hidden layers of the MLP in $P$ as 2, and $\lambda_{pri}=0.1$, $\lambda_{con}=1$, $\lambda_{intra}=1$, $\lambda_{inter}=1$, $\lambda_{reg}=20$, and learning rate as $3\times 10^{-4}$. We set the identity indicator $l$ in the cosine embedding loss as 0.4. We train the model for 20 epochs with the Adam optimizer \cite{kingma2017adam}, where $\beta_1 =0.9$ and $\beta_2=0.999$. The training process takes only 1 hour on a single NVIDIA GTX 1080Ti GPU.


To demonstrate the generality of our framework, instead of using the FaceNet (which is used for training) as the face recognizer, we use a different architecture for recognizing the virtual face images in testing. Unless otherwise stated, we use the SphereFace \cite{liu2017sphereface} which is pre-trained on CASIA-WebFace \cite{yi2014learning} as the face recognizer for evaluating the performance of our virtual face images.

\begin{figure*}[t]
    \centering
    \includegraphics[width=0.9\textwidth]{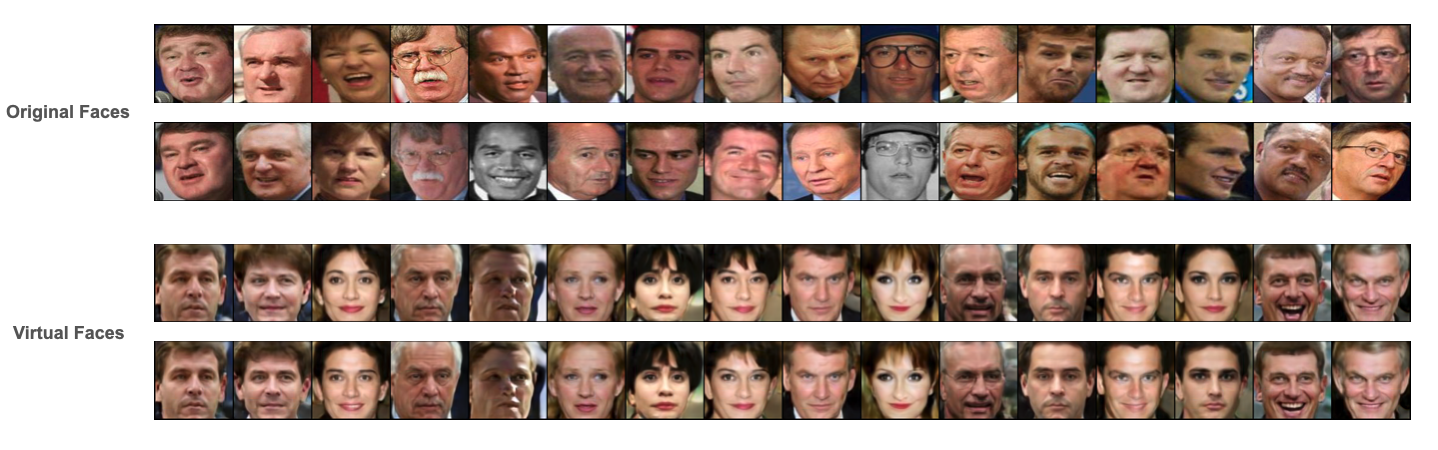}
    \caption{Examples of our virtual face images generated from the LFW test set. Images in first two rows are the original faces images, where the corresponding virtual face images are given in the third and fourth rows.}
    \label{figure4}
\end{figure*}

\subsection{Performance}
To the best of our knowledge, we are the first to propose and formalize the task of generating identifiable virtual face images. Here, we compare the performance of our method with the most relevant work proposed by Gu \emph{et al.} \cite{gu2020password}. Both our scheme and the work in \cite{gu2020password} use a key and an original face image for input, which are able to generate real-look alike virtual/encrypted face images using different keys.

\begin{table}[t]
    \centering
     \caption{Face recognition performance on different image sets. }
    \begin{tabular}{lcc}
        \toprule
        Face image sets & EER & AUC \\
        \midrule
        Original & 0.024 & 0.998 \\
        \midrule
        Gu \emph{et al.} \cite{gu2020password} (Set-A) & 0.105$\pm$0.009 & 0.957$\pm$0.004 \\
        Gu \emph{et al.} \cite{gu2020password} (Set-B) & 0.136$\pm$0.012 & 0.929$\pm$0.006 \\
        Ours (Set-A) & \textbf{0.018$\pm$0.005} & \textbf{0.999$\pm$0.001} \\
        Ours (Set-B) & 0.103$\pm$0.009 & 0.963$\pm$0.004 \\
        \bottomrule
    \end{tabular}
    
    \label{table1}
\end{table}

        
    

\begin{table}[t]
    \centering
    \caption{Face recognition performance of our virtual face images generated from CelebA with ArcFace model as the face recognizer. }
    \begin{tabular}{lcc}
        \toprule
        Face image sets & EER & AUC \\
        \midrule
        Original & 0.048 & 0.975 \\
        \midrule
        Set-A-CelebA & \textbf{0.035$\pm$0.002} & \textbf{0.994$\pm$0.001} \\
        Set-B-CelebA & 0.196$\pm$0.004 & 0.887$\pm$0.003 \\
        
        \bottomrule
    \end{tabular}
    
    \label{table1celeba}
\end{table}

\subsubsection{Face Recognition}
In this section, we evaluate the performance of our virtual faces in terms of face recognition, including Equal Error Rate (EER) and Area Under Curve (AUC). We generate the following two sets of virtual face images from the LFW test set:
\begin{itemize}
\item Set-A: We randomly assign a key to each identity to generate the virtual face images.
\item Set-B: Each identity is assigned with the same key to generate the virtual face images.
\end{itemize}
Set-A refers to the case that the user's key is not known by the adversary, while Set-B is generated for evaluating the key stolen case. We generate the same sets of encrypted face images using the method proposed in \cite{gu2020password} for comparison. Table \ref{table1} gives the EER and AUC among different sets of face images, where ``original" refers to the set of the original face images. It can be seen that, our virtual face images achieve the best recognition performance when the key is well protected, which is even better than that of the original face images. When the key is stolen, the EER drops around 8\% using our virtual face images.


Next, we evaluate our method on the CelebA test set using another face recognizer which is the pre-trained Arcface model \cite{deng2019arcface}. We test on two sets of virtual face images as mentioned before, including Set-A-CelebA and Set-B-CelebA, for virtual face images generate using random keys and the same key for different identities. The results are shown in Table\ref{table1celeba}. Again, when the keys are protected (i.e., Set-A-CelebA), our virtual face images offers higher recognition accuracy compared with the original face images. This indicates that our scheme works on arbitrary face images and can be recognized using arbitrary face recognizers.

Our IVFG supports different key spaces according to the requirements of different applications. For a different key space, the only thing we have to do is to make the dimension of the input of the MLP (which is in the latent projector $P$ of the IVFG) to be the same as the dimension of the combination of the facial representation $r$ and the key $k$. We adjust and retrain our IVFG on a 128-bit key space with $k \in \{0, 1\}^{128}$, where we are still able to generate identifiable virtual face images. The EER of the virtual face recognition on the LFW test set is 0.016$\pm$0.005 and 0.112$\pm$0.013 for the key protected and key stolen cases, respectively. These are similar to the results of using the default 8-bit keys (see Table \ref{table1}). Thus, our IVFG is flexible and robust against different key spaces.

\begin{figure*}[t]
    \centering
    \includegraphics[width=0.9\textwidth]{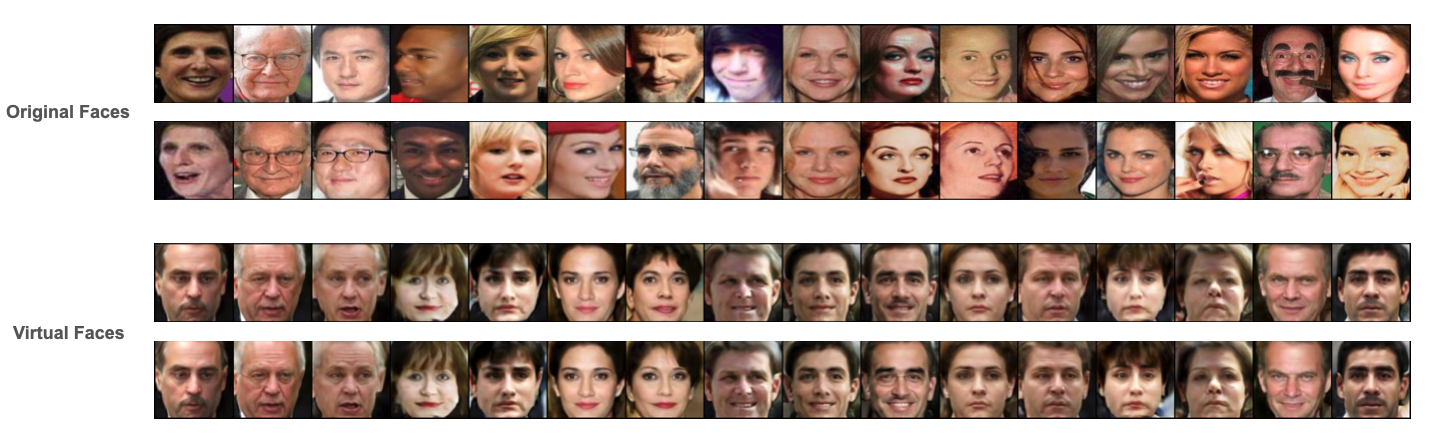}
    \caption{Examples of our virtual face images generated from the CelebA test set. Images in first two rows are the original faces images, where the corresponding virtual face images are given in the third and fourth rows.}
    \label{figure4celeba}
\end{figure*}

\subsubsection{Anonymization}
We evaluate the anonymization performance of our virtual face images in terms of protection rate, which is computed as the unsuccessful match rate between the original and anonymized face images. Specifically, we match the original face image and the corresponding anonymized face image. If the match score is less than the EER threshold of the face recognizer, these two images are considered to be from different identities and an unsuccessful match is yielded. Table \ref{table3} reports the protection rate of our virtual face images and Gu \emph{et al.} \cite{gu2020password}'s encrypted face images on the test set of LFW. It can be seen that our virtual face images offer excellent anonymization with protection rate of 0.988, which is significantly better than Gu \emph{et al.}'s method.

\begin{figure*}[t]
    \centering
    \includegraphics[width=0.8\textwidth]{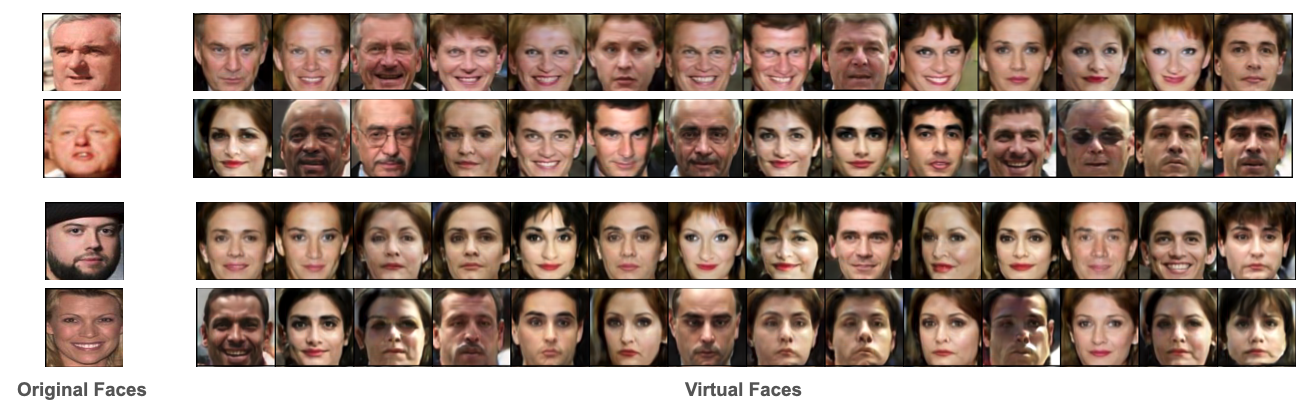}
    \caption{Diverse virtual face images generated from the same face image with different keys. In each row, the first image is the original face images, the rest images are the corresponding virtual face images generated using different keys. The original face images of the first two rows are randomly selected from the LFW test set, while those of the last two rows are randomly selected from the CelebA test set.  }
    \label{figure5}
\end{figure*}

\subsubsection{Diversity}
We evaluate here how different virtual face images could be generated using different keys. For each face image from the test set of LFW, we use two different keys to generate two virtual face images. We match these two virtual face images to see if the matching is successful (i.e., the matching score is over the EER threshold). The diversity is measured as the unsuccessful match rate among the virtual face images generated using different keys, the results of which are given in Table \ref{table3}. It can be seen that our method performs significantly better than the work in \cite{gu2020password} in terms of diversity.

\begin{table}[t]
    \centering
    \caption{Performance on anonymization and diversity.}
    \begin{tabular}{lcc}
        \toprule
        Metrics & Gu \emph{et al.} \cite{gu2020password} & Ours\\
        \midrule
        Protection Rate & 0.052$\pm$0.007 & \textbf{0.988$\pm$0.005} \\
        Diversity & 0.008$\pm$0.003 & \textbf{0.750$\pm$0.013} \\
        \bottomrule
    \end{tabular}
    
    \label{table3}
\end{table}

\begin{table}[t]
    \centering
    \caption{Detection rate and FID scores between the original and anonymized faces. The scores are calculated on features from the second max pooling layer with a dimension of 192.}
    \begin{tabular}{lcc}
        \toprule
        Methods & Detection & FID \\
        \midrule
        Pixelization & 0.984 & 44.451 \\
        Blurring & 0.991 & 62.786 \\
        Gu \emph{et al.} \cite{gu2020password} & 1.0$\pm$0.001 & \textbf{4.782$\pm$0.027}  \\
        Ours & \textbf{1.0$\pm$0.0} & 6.170$\pm$0.050 \\
        \bottomrule
    \end{tabular}
    
    \label{table5}
\end{table}

\subsubsection{Recoverability}
Next, we evaluate the possibility for the attackers to recover the original face image from the virtual face images when the key is stolen. In particular, we use the virtual face images in Set A and the corresponding keys for input and run the IVFG again to generate a set of recovered face images. We match these recovered face images with the original face images, the successful match rate of which is 0$\pm$0.001 at the EER threshold. This is to say, it is difficult to recover the original face images when our virtual face image, the key and the trained IVFG are compromised. While in Gu \emph{et al.} \cite{gu2020password}'s work, the original face image can be directly decrypted from the encrypted face image using the correct password.

\begin{table*}[t]
    \centering
     \caption{Ablation study of our model. From the second row to the sixth row, each row shows the performance of our virtual faces with one loss component removed.}
    \begin{tabular}{lcccccccc}
    \toprule
    Methods & Protection Rate & Diversity & EER (same key) & EER (different keys) & FID \\
    \midrule
    w/o $\mathcal{L}_{pri}$   & 0.978$\pm$0.006      & 0.718$\pm$0.010 & 0.105$\pm$0.008 & 0.021$\pm$0.009 & 6.066$\pm$0.056 \\
    w/o $\mathcal{L}_{con}$   & 0.987$\pm$0.003      & 0.0$\pm$0.0   & 0.108$\pm$0.005 & 0.096$\pm$0.004 & 7.641$\pm$0.008 \\
    w/o $\mathcal{L}_{intra}$ & 0.968$\pm$0.008  & 0.764$\pm$0.019 & 0.190$\pm$0.008 & 0.112$\pm$0.008 & 5.776$\pm$0.086 \\
    w/o $\mathcal{L}_{inter}$ & 0.982$\pm$0.004      & 0.803$\pm$0.015 & 0.180$\pm$0.012 & 0.0$\pm$0.0 & 6.339$\pm$0.037 \\
    w/o $\mathcal{L}_{reg}$   & 0.996$\pm$0.002      & 0.740$\pm$0.021 & 0.096$\pm$0.007 & 0.016$\pm$0.005 & 15.535$\pm$0.144 \\
    \midrule
    full objective& 0.988$\pm$0.005      & 0.750$\pm$0.013 & 0.103$\pm$0.009 & 0.018$\pm$0.005 & 6.170$\pm$0.050 \\
    \bottomrule
    \end{tabular}
   
    \label{table4}
\end{table*}

\begin{figure*}[t]
    \centering
    \includegraphics[width=0.9\textwidth]{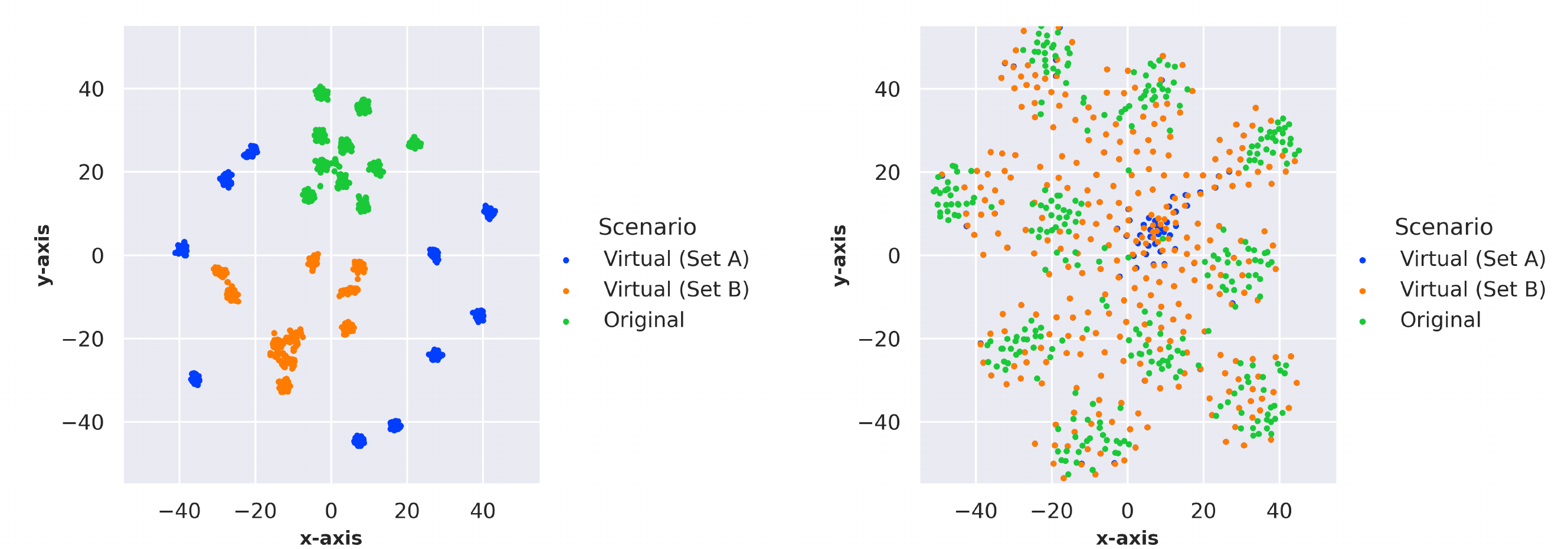}
    \caption{t-SNE visualization of the feature space of different face image sets, left: ours, right:  Gu \emph{et al.} \cite{gu2020password}'s }
    \label{fig6}
\end{figure*}

\subsubsection{Visual Quality}
To evaluate the visual quality of the virtual face images, we report the detection rate using the face detector MTCNN \cite{zhang2016joint} as well as the Fr\'{e}chet Inception Distance (FID) \cite{heusel2018gans} to quantify the similarity of data distributions between the virtual face images and real ones. The face detection thresholds using MTCNN are set as the default values. Table \ref{table5} gives the results for our proposed method, Gu \emph{et al.}'s method and traditional methods including pixelization (with pixel size of 18) and blurring (Gaussian blur with kernel size of 37 and $\sigma$=6). Our method achieves perfect detection rate of 1.0. In terms of FID, our method significantly outperforms traditional anonymization methods and achieves comparable results with the work in \cite{gu2020password}.

Fig. \ref{figure4} and Fig. \ref{figure4celeba} give a few examples of our virtual face images as well as the corresponding original ones from the LFW and CelebA test sets, respectively. In these two figures, the four face images in each column belong to the same identity, and each identity is bound with a user specific key for virtual face generation. It can be seen that our virtual face images are visually real-look alike, which are completely different from the original ones. On the other hand, the virtual face images generated from the face images of the same identity are similar. This is because they are generated from the user specific key for the same identity, and the outputs of our network are highly bound with the identity and key.     

Fig. \ref{figure5} further illustrates the diverse virtual face images that are generated from the same face image with different keys. We can see that, by changing the keys, we are able to generate virtual face images with diverse visual appearances. Therefore, when one of the virtual face images is compromised, it can be easily replaced with another using a different key.

\subsection{Feature Space Visualization}
In this section, we visualize the feature space of the original and virtual face image for qualitative analysis. We randomly select 10 identities from the LFW test set with at least 30 face images, from which we generate different sets of virtual face images using different keys (i.e., Set A) or the same key (i.e., Set B). We perform t-SNE on the extracted facial features of the original and virtual faces \cite{van2008visualizing} in Set A and Set B, the visualization results of which are shown in Fig. \ref{fig6}. We can see that, in the case of different keys (i.e., Set A), the virtual facial features are more sparsely distributed than the original ones, which can be well classified in the virtual space. On the other hand, the features of our virtual face images can also be well separated from the original ones, leading to good anonymization ability. When the key is stolen (Set B), the virtual features are still clustered to a certain extent. We also compare our virtual feature space with that of Gu \emph{et al.}'s. It can be seen that, by using Gu \emph{et al.}'s method, the features of the encrypted face images are not well clustered, which are also difficult to be differentiated from the original ones.

\subsection{Ablation Study}
In this section, we present ablation studies to demonstrate the effectiveness of each loss component in our multi-task learning objective on the test set of LFW. We remove each loss component from the overall training objective separately and report the results in Table \ref{table4}.

It can be seen from Table \ref{table4} that the $\mathcal{L}_{pri}$ effectively pushes the features apart, the protection rate drops if $\mathcal{L}_{pri}$ is removed. Moreover, the average cosine similarity drops from 0.0052 to 0.0024 between virtual and original facial features when $\mathcal{L}_{pri}$ is incorporated during training. Without $\mathcal{L}_{con}$, the model fails to be conditioned on the user-specified key to generate different virtual face images. The results of w/o $\mathcal{L}_{intra}$ and w/o $\mathcal{L}_{inter}$ demonstrate that both the intra-group and inter-group classification losses are necessary for high face recognition accuracy. The results of w/o $\mathcal{L}_{reg}$ shows that $\mathcal{L}_{reg}$ regularizes the anonymization process to produce realistic face images.

\section{Conclusions}
In this paper, we propose and formalize a novel task in face anonymization: generating identifiable virtual face images.
To this end, we propose a framework termed as the IVFG for the generation of identifiable virtual face images. The input of the IVFG is the original face image and a user specific key, the output is a virtual face image bound with a unique virtual identity. Our virtual face images are visually different from the original ones for protecting the original identity, which could also be directly recognized in the virtual space for authentication. We also propose a multi-task learning objective and a triplet styled training strategy to effectively train the IVFG. Experimental results demonstrate that our virtual face images achieve good performance for face anonymization and identification of virtual identities.

\begin{acks}
This work was supported in part by the National Natural Science Foundation of China under Grant 62072114, U20A20178,U20B2051 and U1936214, in part by Alibaba Group through Alibaba Innovative Research (AIR) Program.
\end{acks}

\bibliographystyle{ACM-Reference-Format}
\balance 
\bibliography{ref}
\end{document}